# Detecting mental disorder on social media: a ChatGPT-augmented explainable approach

Loris Belcastro[a,*], Riccardo Cantini[a], Fabrizio Marozzo[a], Domenico Talia[a], Paolo Trunfio[a]

[a]*DIMES - University of Calabria, Ponte P. Bucci 41/C, Rende, 87036, , Italy*

---

## Abstract

In the digital era, the prevalence of depressive symptoms expressed on social media has raised serious concerns, necessitating advanced methodologies for timely detection. This paper addresses the challenge of interpretable depression detection by proposing a novel methodology that effectively combines Large Language Models (LLMs) with eXplainable Artificial Intelligence (XAI) and conversational agents like ChatGPT. In our methodology, explanations are achieved by integrating BERTweet, a Twitter-specific variant of BERT, into a novel self-explanatory model, namely BERT-XDD, capable of providing both classification and explanations via masked attention. The interpretability is further enhanced using ChatGPT to transform technical explanations into human-readable commentaries. By introducing an effective and modular approach for interpretable depression detection, our methodology can contribute to the development of socially responsible digital platforms, fostering early intervention and support for mental health challenges under the guidance of qualified healthcare professionals.

*Keywords:* XAI, Natural Language Processing, Large Language Models, BERT, ChatGPT, Mental disorder, Depression detection

---

*Corresponding author

*Email addresses:* lbelcastro@dimes.unical.it (Loris Belcastro), rcantini@dimes.unical.it (Riccardo Cantini), fmarozzo@dimes.unical.it (Fabrizio Marozzo), talia@dimes.unical.it (Domenico Talia), trunfio@dimes.unical.it (Paolo Trunfio)

## 1. Introduction

In the era of social media, people openly share their emotions, thoughts, and experiences, generating vast amounts of digital data that reflect their opinions, emotions, and mental states. As a result, social media has become a crucial tool across various fields, serving not only as a platform for communication and information sharing but also as a valuable resource for detecting issues and monitoring societal trends, including mental health, public sentiment, political polarization, and broader social challenges [1, 2, 3]. Notably, some users post content that may indicate mental health disorders or depressive symptoms, highlighting the need for effective detection mechanisms. However, the sheer volume and dynamic nature of user-generated content pose significant challenges for accurate and timely identification.

Over the years, Natural Language Processing (NLP) models based on the Transformer [4] architecture, such as *BERT (Bidirectional Encoder Representations from Transformers)* [5] and *GPT (Generative Pre-trained Transformer)* [6], have gained prominence due to their ability to capture contextual nuances in language. Besides their sophisticated architecture, the effectiveness of these models is also due to the training on large and heterogeneous datasets gathered from several sources, carried out on high-performance and distributed architectures [7, 8]. Nonetheless, these models face challenges in providing comprehensible explanations for their predictions, as they inherently lack interpretability, hindering their application in critical domains such as mental health diagnosis [9]. To tackle these challenges, XAI techniques have emerged, which encompass *post-hoc* and *self-explanatory* techniques [10]. Post-hoc techniques aim to explain predictions from pre-trained black-box models. Currently, the most popular approaches are *model-agnostic*, meaning they can be applied to any underlying black-box model, with no assumption on their internal working and structure. Among them, LIME (Local Interpretable Model-agnostic Explanation) [11] and SHAP (SHapley Additive exPlanations) [12] determine a weight assignment as a proxy for feature importance by following a regression and game theory approach, respectively. Differently, self-explanatory techniques are trained to provide explanations alongside predictions. However, these methods commonly encounter challenges related to flexibility and integration with other deep learning models [13, 14].

Starting from the above consideration, our work addresses the need for an efficient and interpretable approach to detect signs of depression in social media posts, shedding light on the specific words influencing mental status identification and classification. These explanations can be crucial in supporting clinicians, enabling them to understand the rationale behind the model's predictions and facilitating precise and timely diagnosis. In particular, this work introduces an explainable approach for depression classification from social data, which effectively combines transfer learning from large language models (LLMs) [15], explainable artificial intelligence (XAI) [10], and the natural language generation capabilities of conversational agents like ChatGPT. Specifically, a BERT-like LLM is fine-tuned on a corpus of social media posts to make it able to distinguish subtle linguistic patterns indicative of depressive states. Afterward, XAI techniques are leveraged to provide users with comprehensible insights into the features influencing the decision-making process of the model. Finally, the interpretability of the results is enhanced by leveraging ChatGPT to convert the technical explanations provided by the model into a more user-friendly format, i.e., a concise and human-readable text.

The modular nature of the proposed methodology facilitates full customization, giving the possibility to combine different models tailored to specific use cases. Indeed, the methodology seamlessly integrates with both post-hoc XAI techniques. In addition, any classification model and any alternative to ChatGPT can be used, like Meta's Llama or Google Bard. In this work, we leveraged the BERTweet model [16], a Twitter-specific variant of BERT that effectively handles features such as hashtags, mentions, URLs, and emojis. BERTweet has demonstrated superior performance on tweet-based NLP tasks, outperforming state-of-the-art models like RoBERTa and XLM-R [17] across various social media datasets. Explanations are achieved by integrating BERTweet into a novel self-explainable model, namely *BERT-XDD (BERT-based eXplainable Depression Detection)*, which combines it with a bi-directional Long Short-Term Memory (LSTM), generating both a classification and an explanation via masked attention. Finally, ChatGPT is used to transform model explanations into easily understandable human-readable texts via prompting.

The contributions of our work, whose effectiveness was assessed on two datasets encompassing social posts from Twitter and Reddit, two widely used social media platforms, are multifold:

- We propose a modular methodology that allows the creation of an effective pipeline for explainable depression detection from social media

data, considering factors such as the nature of the data and the explanation requisite for end users.

- We introduce a novel self-explainable model that simultaneously outputs a classification and the corresponding explanation, based on the combination of BERTweet and a bi-directional LSTM enhanced with masked attention, effectively balancing accuracy with interpretability.
- The work bridges the gap between complex model explanations and user comprehension, making the explanation more accessible by expressing it into a human-readable format through one-shot prompting with ChatGPT.

Overall, the proposed methodology contributes to enhancing the accuracy, trustworthiness, and applicability of mental health detection systems in the digital age, fostering the development of more socially responsible digital platforms by facilitating early intervention and support for individuals experiencing mental health challenges. In addition, it is worth noticing that our methodology is intended for use by qualified healthcare professionals, including doctors and psychologists, who can validate and leverage the results in their clinical assessments.

The structure of this paper is as follows. Section 2 provides an overview on existing research in the field of depression detection through social data analysis. Section 3 describes the proposed methodology. Section 4 discusses the achieved results. Finally, Section 5 concludes the paper.

## 2. Related work

Research on detecting depression and mental disorders through textual and linguistic analysis is extensive. Studies primarily focus on exploring correlations between mental health symptoms and language to identify disorders like depression, suicidal tendencies, and other health conditions (e.g., flu, pregnancy, and eating disorders) [18]. The exponential growth of available data, also fostered by the widespread diffusion of social media, has led to the development of data-driven solutions that leverage machine learning and deep learning techniques for the automated detection of mental disorders [19, 20, 21]. Several works have been proposed in the literature for detecting depression signals from social media data using machine learning techniques. These works encompass a wide range of algorithms, including multi-kernel SVM, Naive Bayes, ensemble approaches like Random Forest, and emotion-based dictionaries [22, 23, 24, 25, 26, 27]

In recent years, deep learning-based has surpassed other machine learning techniques in various domains, offering the advantage of automatic feature extraction without complex operations. Because of this, have been conducted numerous studies using deep neural networks to identify mental disorders in social media users. Tejaswini et al. [28] addressed challenges in early depression detection using social media data combining NLP techniques with a hybrid learning approach integrating fastText embeddings, CNN, and LSTM architectures. Chen et al. [29] introduced the hybrid deep learning model SBERT-CNN for detecting depression on social media. They combined pre-trained sentence BERT to capture semantic information with CNNs, which identify temporal behavioral patterns in users. Wang et al. [30] explored transformer-based architectures, including BERT, RoBERTa, and XLNet, for predicting depression risk at four levels using Weibo data. Poświata and Perełkiewicz [31] compared different transformer-based models, with RoBERTa emerging as the best performer. The authors also fine-tuned their own language model, namely DepRoBERTa (RoBERTa for Depression Detection), which was previously pre-trained on depression-related posts from Reddit. Finally, to enhance detection accuracy, they ensemble-averaged the DepRoBERTa model with a fine-tuned RoBERTa-large model. Recent techniques have also emerged in the field of depression detection that leverage XAI approaches. For example, Zogan et al. [32] introduced a depression detection model, based on a bidirectional gated recurrent unit-convolutional neural network (BiGRU-CNN), designed to tackle the issue of model interpretability when analyzing social media data, by incorporating a hierarchical attention mechanism.

Unlike other works that focus solely on performance or interpretability, BERT-XDD effectively combines both by introducing a novel self-explainable model capable of providing explanations alongside high classification performance. To this end, our methodology leverages BERTweet, designed for handling social media data effectively and reliably, and combines it with a bi-directional LSTM enhanced with masked attention. Additionally, BERT-XDD improves explanations by translating them into a human-readable format through ChatGPT. This ensures that the model decisions are more transparent and understandable to end users, which is particularly crucial in clinical settings where interpretability significantly impacts trust and usability.

## 3. Proposed methodology

The proposed methodology provides an effective and interpretable approach to identify signs of depression in user-generated content on social media. It relies on a modular architecture that effectively combines: *i*) the use of large-scale pre-trained LLMs for depression classification via transfer learning; *ii*) XAI techniques to obtain useful insights into the rationale behind the model classification process, in terms of word-level feature importance; *iii*) conversational AI to transform the explanations provided by the model into a human-readable format, ensuring higher accessibility for clinicians and users. In the following, we provide a detailed description of the main steps of our approach, whose execution flow is depicted in Figure 1.

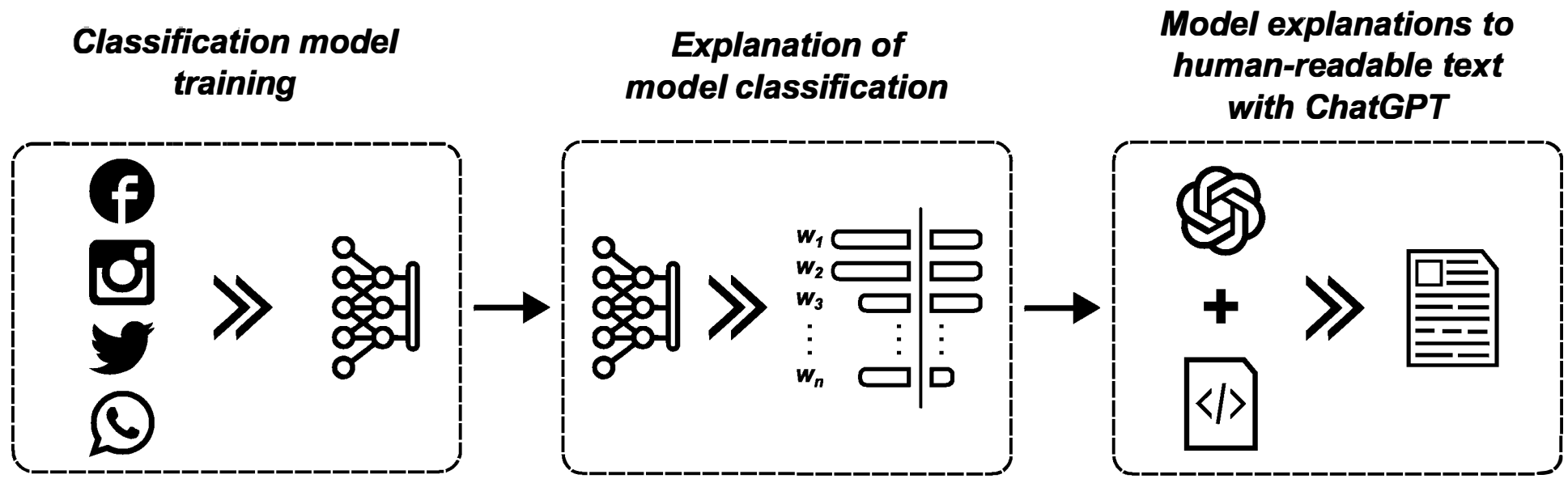

Figure 1: Overview of the BERT-XDD model architecture for detecting and interpreting signs of depression on social media.

### *3.1. Training of the self-explainable classification model*

The novel self-explainable model for depression detection BERT-XDD is used within the proposed pipeline for achieving both classifications and explanations. BERT-XDD combines the BERTweet model with an attention-based bi-directional LSTM. As detailed in the following, the model undergoes a two-phase training approach: initially, a preliminary fine-tuning of the BERTweet classifier is performed on a corpus of social media posts for the downstream task of depression classification. Then the fine-tuned model is integrated and jointly optimized with an attention-based bi-directional LSTM, which is specifically designed to produce an output explanation via masked attention. In particular, our approach assigns weights to individual input words in proportion to their significance for the model's classification, while excluding irrelevant words from the perspective of explanation.

### 3.1.1. Pre-fine-tuning of BERTweet

In this step, the BERTweet model is fine-tuned for the depression detection task. The model is composed of the transformer encoder and a classification head. Given an input sentence $x$, composed by a sequence of words $w_1, w_2, \ldots, w_k$, the encoder computes an embedded representation of $x$ via the self-attention mechanism. This representation consists of a sequence of $k$ $d$-dimensional vectors $E = e_1, e_2, \ldots, e_k$, with $d = 1024$, where $e_i \in \Re^d$ is the embedding of $w_i$. Besides the embedding of the different words, the encoder computes the hidden state associated with the $CLS$ special token (i.e., $e_{CLS}$), which condenses the information of the whole sequence and is used for the downstream classification task. Specifically $e_{CLS}$ is fed as input to the classification head, which computes the final classification as a softmax distribution on a vector of logits $l \in \Re^{|\mathcal{C}|}$, where $\mathcal{C}$ is the set of output classes. The vector $l$ is computed as follows:

$$p_{out} = tanh(W_p \cdot e_{CLS} + b_p) \ , \quad l = W_l \cdot p_{out} + b_l \tag{1}$$

Firstly, the pooler output $p_{out} \in \Re^d$ is computed via a fully connected (FC) layer equipped with a tanh activation, where $W_p \in \Re^{d \times d}$ and $b_p \in \Re^d$ are learnable weights. Afterward, the pooler output $p_{out}$ is fed to a linear FC layer to compute the vector of logits $l \in \Re^{|\mathcal{C}|}$, where $W_p \in \Re^{|\mathcal{C}| \times d}$ and $b_p \in \Re^{|\mathcal{C}|}$ are learnable weights. The model is fine-tuned for 8 epochs by using a batch size equal to 16 and a cross-entropy loss, selecting the best model in terms of macro-F1 score on the validation set. We set the length of the sequence $k$ equal to 200 and used the Rectified Adam [33] optimizer with a learning rate equal to $3 \cdot 10^{-5}$: such a small value is crucial in this phase as we only want to slightly refine the pre-trained weights of BERTweet to work with our depression classification task, and therefore large weight updates are not desirable, due to the risk of overfitting and catastrophic forgetting.

### 3.1.2. BERT-XDD model training

Once the BERTweet model is fine-tuned, as described in the previous step, it is used to build the BERT-XDD model. Specifically, only the encoder of BERTweet is retained, while the classification head is replaced with an attention-based bi-directional LSTM. Given an input sequence $x$ of length $k$, the BERTweet encoder produces a $d$-dimensional embedding for each element in $x$, generating the embedding matrix $E \in \Re^{d \times k}$. This matrix is fed to the bi-LSTM layer that produces a list of hidden representations $h_1, h_2, \ldots, h_k$, i.e., the matrix $H \in \Re^{2u \times k}$. Each hidden representation $h_i$ is represented by the concatenation of left-to-right and right-to-left representations, i.e., $h_i = \overrightarrow{h_i} | \overleftarrow{h_i}$, where both are $u$-dimensional vectors. Afterward, a weight for each hidden state in $H$ is computed via a Bahdanau-like attention mechanism [34] with masking. In particular, a score vector $\sigma$ is computed that measures the unnormalized importance of each of the $k$ elements of the sequence for the model classification. This vector is computed via a parameterized feed-forward neural network as follows:

$$\sigma = v^T tanh(U \cdot H) \tag{2}$$

The learnable matrix $U \in \Re^{2u \times 2u}$ is used to determine a linear projection of $H$ that is fed to a tanh layer. Afterward, a learnable vector $v \in \Re^{2u}$ is used to compute the final vector $\sigma \in \Re^k$. In addition, an attention mask is used to prevent the model from focusing on irrelevant words from the perspective of explanation. In particular, for each element of the sequence, a mask vector $\mu \in \Re^k$ is computed as follows: $\mu_i = 0$ if $w_i$ is a stopword, punctuation, or a special token like the $CLS$ or $PAD$; $\mu_i = 1$ otherwise. In this way, we force the model to only attend to real and meaningful words in the input sequence. The mask is applied to $\sigma$ and the final attention weights $\alpha \in \Re^k$ are computed using a softmax function as follows:

$$\mu_i = (\mu_i - 1) \cdot 10^4 \ , \quad \sigma_i = \sigma_i + \mu_i \tag{3}$$

$$\alpha_i = \frac{e^{\sigma_i}}{\sum_{j=1}^{k} e^{\sigma_j}} \tag{4}$$

In the score calculation, the mask is modified such that values masked with 0 are pushed to very negative numbers, while values masked with 1 remain untouched. In this way, when $\alpha$ weights are computed via the softmax distribution (equation 4), values masked with 0 will be given an exponentially small value, preventing the model from attending to them. Finally, once the masked attention weights are computed, a weighted average of the BERTweet embeddings $E$ is computed, and the resulting vector $\hat{e} \in \Re^d$ is fed to a softmax layer to determine the output class:

$$\hat{e}_i = \sum_{j=1}^{k} E_{ij} \cdot \alpha_j \ , \quad \pi = softmax(W_{out} \cdot \hat{e} + b_{out}) \tag{5}$$

where $W_{out} \in \Re^{\mathcal{C} \times d}$ and $b_{out} \in \Re^{\mathcal{C}}$ are learnable weights. The model outputs the predicted class $c$ for the input sequence $x$ as $c = argmax(\pi)$, and an explanation $\mathcal{E}(x)$. Specifically, the explanation consists of a list of pairs $(w_i : \alpha_i)$ indicating a weight $\alpha_i$ for each word $w_i$, which measures how much attention the model paid to $w_i$ in classifying $x$ as belonging to class $c \in \mathcal{C}$. The BERT-XDD model is trained following a two-step process:

1. *Training of the bi-LSTM head*: in this step, the BERTweet encoder is freezed, and only the bi-LSTM parameters and masked attention matrices are updated. The model is trained for 6 epochs using the ADAM optimizer, initialized with the default learning rate $10^{-3}$.
2. *End-to-end fine-tuning*: in this step the BERTweet encoder is unfreezed and the whole model is fine-tuned end-to-end, for 2 epochs using the same setting described in the pre-fine-tuning step.

### *3.2. Enhancing model explanations with ChatGPT*

As described in the previous section, given an input sequence $x$ composed of a list of words $w_1, w_2, \ldots, w_k$, the self-explainable classification model BERT-XDD outputs a class label $c \in \mathcal{C}$ alongside an explanation $\mathcal{E}(x)$ indicating the influence of each word in $x$ on the model's classification. At this point, we utilize ChatGPT to add a natural language layer to the provided explanation. In practice, ChatGPT generates a brief yet insightful commentary on the classification, starting from the input post $x$, the predicted class $c$, and the explanation $\mathcal{E}(x)$, via prompting.

As illustrated in Figure 2, the prompt has been defined through an incremental prompt engineering process:

1. First, we defined a *base prompt*, which takes as input the social media post $x$, the output class $c$, the explanation $\mathcal{E}$ (generated by the model as a list of words and their relative weights), and a task instructing ChatGPT to generate a brief commentary discussing the assignment of the classification class according to the provided explanation. Note that the prompt shown in Figure 2 refers to the Twitter-based case study (see Section 4), which focuses on the three-way depression classification task with classes `SEVERELY_DEPRESSED`, `MODERATELY_DEPRESSED`, and `NOT_DEPRESSED`.
2. Next, we extended the base prompt to obtain an improved version, namely *advanced prompt*, which introduces some constraints forcing ChatGPT to limit the structure and length of the commentary, also

***advanced prompt***

***base prompt***

**Post (`$x`)**: {*input post x*}

**Model classification (`$c`)**: {*output class c*}

**Explanation (`$e`)**: {(*word : importance*) *pairs in* $\mathcal{E}(x)$}

**Task**: You are given a post `$x` written by a social media user and its classification `$c` as either "`SEVERELY_DEPRESSED`", "`MODERATELY_DEPRESSED`", or "`NOT_DEPRESSED`". Alongside the classification, an explanation `$e` is provided in the form of a list of (word: weight) pairs sorted according to their relevance in descending order, indicating the impact of each word on the model classification. Show me the explanation `$e` as a brief commentary, reporting the explanation's words in quotes.

Do not introduce any information not present in `$e`, and do not add your own comments or deductions. Commentary must be 100 words or fewer and not use bullet points.

Use the following JSON as an example:

{
  `$x`: {*example post* $x'$}, `$c`: {*output class* $c'$}, `$e`: {(*word : importance*) *pairs in* $\mathcal{E}(x')$},
  `commentary`: {*example commentary*}
}

Figure 2: Prompts (basic and advanced) used to enhance the interpretability of the model classifications with explanations generated by ChatGPT.

avoiding its own comments and deductions. Moreover, the *advanced prompt* leverages a one-shot in-context learning approach, which consists of showing the model one JSON example of how the task should be performed, by providing both the input and the desired output. In particular, in the prompt we provide an example post $x'$, the output class $c'$ and corresponding explanation $\mathcal{E}$ produced by BERT-XDD, and an example commentary. We also followed a class-specific strategy, i.e., the input post and the provided example share the same class ($c = c'$).

This approach, relying on injecting examples into the context of LLMs via text interaction, can lead to superior performance with respect to using them in a zero-shot fashion [6]. In our specific case, using one-shot demonstrations helps to ensure that the description generated by ChatGPT adheres strictly to the explanation $\mathcal{E}(x)$ provided by BERT-XDD. This aspect is crucial since, by avoiding the introduction of external information or personal deductions, this *ChatGPT-augmentation* process preserves the explanation transparency, while allowing the user to better interpret the model classification, presenting it in a user-friendly format.

## 4. Experimental results

BERT-XDD combines a fine-tuned BERTweet model with an attention-based bi-directional LSTM to provide a self-explainable approach to depression classification. The model performance was assessed on two datasets of social media posts, labeled for the depression detection task. The first dataset [31] is a preprocessed version of the original one used for the *Shared Task on Detecting Signs of Depression from Social Media Text* competition at LT-EDI-ACL2022 [35]. It encompasses 10 251 tweets, labeled as `SEVERELY_DEPRESSED`, `MODERATELY_DEPRESSED`, and `NOT_DEPRESSED`, which are divided into 6 006, 1 000, and 3 245 instances for training, validation, and testing, respectively. The second dataset contains 7 650 posts from Reddit [1], annotated as `DEPRESSED` and `NOT_DEPRESSED`. Data are balanced across the two classes, and divided into 5 731, 800, and 1 200 instances for training, validation, and testing, respectively. Experiments have been performed using a machine with 64 CPU cores and an Nvidia A30 GPU with 24GB RAM.

In the following, we report the achieved experimental results, focusing on the main aspects of our work. Section 4.1 discusses the choice of BERTweet as the transformer-based model for BERT-XDD. Section 4.2 presents a performance comparison in terms of classification accuracy with other state-of-the-art models. Finally, Section 4.2.1 shows some examples of ChatGPT-augmented explanations, stressing the importance of prompt engineering and one-shot demonstrations to ensure transparency.

### *4.1. Model comparison for the pre-fine-tuning step*

Among the main alternative models that can be effectively used for detecting depression on social media, we carried out a comparative evaluation of BERTweet against different BERT-like models:

- *BERT* [5]: it is a pre-trained language representation model based on the transformer architecture.
- *RoBERTa* [36]: it is an improved version of BERT, which removes the Next Sentence Prediction (NSP) task from the pre-training phase and introduces dynamic masking to vary the masked tokens during language modeling.

[1]- Accessed February, 2025

- *ALBERT* [37]: it is a lightweight variant of BERT. It introduces some improvements such as factorized embedding parameterization, and inter-sentence coherence loss, by replacing the Next Sentence Prediction with the Sentence order prediction task.

As shown in Table 1, BERTweet outperformed the other transformer-based models in both datasets, highlighting its superior ability to handle textual data from various social media platforms. Therefore, we have chosen BERTweet as the transformer-based model to be used within BERT-XDD.

| Dataset | Metric | BERT | RoBERTa | ALBERT | BERTweet |
|---|---|---|---|---|---|
| *Twitter depression detection* | Accuracy | 0.653 | 0.646 | 0.654 | **0.664** |
| | Precision | 0.535 | **0.571** | 0.539 | 0.559 |
| | Recall | 0.554 | 0.563 | 0.560 | **0.578** |
| | Macro-F1 | 0.540 | 0.565 | 0.548 | **0.566** |
| *Reddit depression detection* | Accuracy | 0.942 | 0.972 | 0.971 | **0.986** |
| | Precision | 0.915 | 0.972 | 0.968 | **0.986** |
| | Recall | 0.933 | 0.971 | 0.969 | **0.985** |
| | Macro-F1 | 0.921 | 0.972 | 0.968 | **0.985** |

Table 1: Comparison of BERT-like models for the pre-fine-tuning step. For all models, the *large* version was used. Bolded and underlined values indicate the best and second-best models, respectively.

### *4.2. Performance evaluation*

The classification accuracy of BERT-XDD has been assessed through an extensive comparison with the following models for depression detection:

- *BERTweet*: the pre-fine-tuned model for depression detection (see Section 3.1.1).

- *DepRoBERTa* [31]: a RoBERTa-based model pre-trained on the Reddit Mental Health Dataset [38] and fine-tuned on the evaluation dataset (*Twitter* or *Reddit*).

- *DepRoBERTa-ensemble*: it consists of the ensemble averaging of DepRoBERTa with a RoBERTa-large fine-tuned on the evaluation dataset (*Twitter* or *Reddit*).

As shown in Table 2, BERT-XDD achieved the best results in both datasets. It is worth noting that, despite not using any additional data for further pre-training, it still outperforms both the base and ensemble versions of DepRoBERTa. Moreover, unlike other models, BERT-XDD can provide a local explanation alongside the classification output, effectively addressing the trade-off between interpretability and accuracy [39]. This balance enables BERT-XDD to offer insight into the rationale behind the model's decision process while maintaining high classification performance, making it a valuable tool for practical applications requiring transparent decision-making.

| **Dataset** | **Metric** | **DepRoBERTa** | **DepRoBERTa ensemble** | **BERTweet** | **BERT-XDD** |
|---|---|---|---|---|---|
| *Twitter depression detection* | Accuracy | 0.626 | 0.658 | 0.664 | **0.664** |
| | Precision | 0.575 | **0.586** | 0.559 | 0.582 |
| | Recall | 0.588 | 0.591 | 0.578 | **0.601** |
| | Macro-F1 | 0.571 | 0.583 | 0.566 | **0.590** |
| *Reddit depression detection* | Accuracy | 0.987 | 0.988 | 0.986 | **0.991** |
| | Precision | 0.987 | 0.988 | 0.986 | **0.991** |
| | Recall | 0.986 | 0.987 | 0.985 | **0.991** |
| | Macro-F1 | 0.987 | 0.988 | 0.985 | **0.991** |

Table 2: Comparison with state-of-the-art depression detection models. Bolded and underlined values indicate the best and second-best models, respectively.

#### *4.2.1. ChatGPT-augmented explanations*

In this section, we show how ChatGPT is employed to make the classification results more understandable to the end user, by presenting the corresponding explanations in the form of natural language text, as described in Section 3. Using the prompts (*base* and *advanced*) defined in Section 3.2, we present some representative examples of commentary obtained for the different classes considered. Furthermore, for the sake of brevity, we only show commentaries generated for the Twitter-based case study, encompassing three different depression levels, namely `SEVERELY_DEPRESSED`, `MODERATELY_DEPRESSED`, and `NOT_DEPRESSED`.

Let's consider a first example of a post that has been correctly classified by our model as `SEVERELY_DEPRESSED`:

> *"Day 19 on antidepressants, it is getting worse again: I am fixated on my failure and that I am alone. I am nearly constantly fatigued, after the few hours I have* *with some energy, it is time to take a tablet again. I think I will ask to try an increased dose.*"

Alongside the predicted class, the BERT-XDD model outputs the following explanation $\mathcal{E}(x)$ indicating the weight of each word in the post on the model's classification:

$\mathcal{E}(x)$ = {"failure": 0.1183, "fatigued": 0.1175, "energy": 0.0931, "time": 0.0904, "fixated": 0.0840, "constantly": 0.0599, "try": 0.0514, "tablet": 0.0504, "think": 0.0431, "nearly": 0.0401, "antidepressants": 0.0401, "hours": 0.0398, "ask": 0.0365, "worse": 0.0362}

Starting from the output of the model, we used ChatGPT to generate a brief yet insightful commentary on the classification via prompting. In the following, we show the results obtained with both the *base* and *advanced* prompt, highlighted in yellow and light blue color, respectively.

**Commentary (*base prompt*):** "The model classifies the given post as `SEVERELY_DEPRESSED` based on the user's depiction of worsening symptoms despite being on "*antidepressants*" for 19 days. The highlighted words in the explanation point to significant indicators of severe depression, including the user's fixation on "*failure*", feelings of loneliness, constant fatigue, and the need for medication to manage their energy levels. The mention of contemplating an increased dose and the struggle with low "*energy*" further accentuate the severity of the user's emotional state. The model's classification aligns with the apparent intensity of the user's depressive symptoms as expressed in the provided text."

**Commentary (*advanced prompt*):** "The model classifies the user of this post as `SEVERELY_DEPRESSED` evident through their fixation on "*failure*" and pervasive feelings of "*aloneness.*" The user describes being "*constantly fatigued,*" with limited periods of "*energy*" followed by the necessity of taking antidepressant tablets. The user contemplates requesting an "*increased dose,*" indicating a challenging treatment journey. The commentary emphasizes the user's struggle with persistent negative thoughts, "*fatigue*", and the need for medication adjustments."

Upon analyzing the commentary generated by ChatGPT, the following patterns of behavior become evident. The *advanced prompt* generates superior commentary compared to the *base prompt*, providing a more concise and focused output. It avoids unnecessary details and redundant explanations, instead highlighting key indicators such as "fixation on failure" and "contemplating increased dose" to succinctly illustrate the severity of the condition. In contrast, the *base prompt* tends to produce outputs that excessively rely on explicit information contained in the original post (e.g., the use of antidepressants for 19 days) while introducing redundant considerations about the model's classification (e.g., the last sentence of the commentary). Overall, the *advanced prompt* consistently performs well and proves to be more reliable, adhering better to the explanation provided by BERT-XDD. These improvements justify the exclusive use of the advanced prompt for future analyses, ensuring results that are both clear and reliable.

Let's now consider the following post, which has been correctly classified as `MODERATELY_DEPRESSED` by our model:

"*I'm sabotaging myself: sleeping doesn't help me, if I stay awake it's easier for me to handle everything because nothings expected of you at night, there's no 'you have to be doing something' other than sleeping, but the night is where I feel most, whether it be good or bad, (mostly bad) I'm just tired. physically and mentally, trying to say the same thing to people who don't listen, trying to explain something but not being able to properly articulate it without getting frustrated. I don't think I can do this anymore, I'm hurting my family without getting help, but I can't get help if I don't want it and am not ready for it yet, what do I do I'm really struggling, this is the worst its ever been before.*"

The explanation provided by the model is the following:

$\mathcal{E}(x)$ = {"sabotaging": 0.1462, "sleeping": 0.0729, "help": 0.0712, "struggling": 0.0647, "worst": 0.0555, "awake": 0.0371, "handle": 0.0346, "stay": 0.0285, "tired": 0.0279, "think": 0.0269, "frustrated": 0.0254, "easier": 0.0245, "anymore": 0.0239, "feel": 0.0215, "night": 0.0199, "bad": 0.0193, "family": 0.0184, "hurting": 0.0169, "getting": 0.0167, "cant": 0.0163, "good": 0.0147, "res": 0.0146, "physically": 0.0143, "mentally": 0.012, "ings": 0.0112, "trying": 0.0105, "expected": 0.0102, "able": 0.0068, "articulate": 0.0068, "want": 0.0065, "ready": 0.0055, "listen": 0.0043, "thing": 0.003, "properly": 0.003, "explain": 0.0019, "people": 0.0014}

The commentary generated by ChatGPT using the *advanced prompt* is as follows.

**Commentary (*advanced prompt*):** "The user's post reveals a deep emotional struggle, pointing towards a MODERATELY_DEPRESSED state. Words such as "*sabotaging,*" "*struggling,*" and "*worst*" convey a sense of internal turmoil. The user describes difficulty articulating their feelings to others and feeling physically and mentally exhausted. The mention of "*hurting*" their family and the inability to seek help due to unreadiness further underscores their distress. The word weights highlight the significance of terms like "*dead,*" "*tired,*" and "*hurting*" in the model's classification. It is crucial to acknowledge the user's pain, encourage open communication, and explore supportive avenues for their well-being."

Finally, let's consider a post classified as NOT_DEPRESSED by our model:

"*Working on myself and have been slowly feeling better, then my fiancé broke off our engagement last night. I'm just kind of lost right now. The break up is super fresh and super complicated. Today is an extremely low day.*"

The explanation provided by the model is the following:

$\mathcal{E}(x)$ = {"low": 0.1416, "day": 0.1190, "lost": 0.1133, "extremely": 0.0871, "today": 0.0811, "right": 0.0669, "kind": 0.0569, "working": 0.0520, "fresh": 0.0390, "better": 0.0386, "complicated": 0.0309, "fiancé": 0.0269, "feeling": 0.0246, "super": 0.0241, "slowly": 0.0224, "engagement": 0.0184, "night": 0.0138, "break": 0.0115, "broke": 0.0099}

The commentary generated by ChatGPT using the *advanced prompt* is as follows.

**Commentary (*advanced prompt*):** "The user's post suggests a state of NOT_DEPRESSED. They express working on themselves and feeling better, but the recent and complex breakup with their "*fiancé*" has left them feeling "*lost.*" The user emphasizes that today is an "*extremely low day.*" The words "*low,*" "*lost,*" and "*extremely*" carry significant weight in the model's classification. The user's acknowledgment of working on themselves and the fresh, complicated nature of the breakup contributes to the overall context. The model recognizes the emotional impact of words like "*fresh,*" "*complicated,*" and "*engagement.*"
"

## 5. Conclusion

Depression affects about 280 million people worldwide, and has been cited as the "*single largest contributor to global disability*" by the World Health Organization. Every day people share on social media digital contents containing valuable insights about their emotions, thoughts, and experiences. This constant flow of information offers a significant opportunity to identify the presence of mental disorders, such as depression, through the usage of advanced detection methodologies. This paper introduces BERT-XDD, a novel self-explainable methodology for detecting signs of depression in social media posts. By combining BERTweet with an attention-based bi-directional LSTM, BERT-XDD not only accurately classifies depression levels but also offers comprehensive explanations. These explanations are then delivered in a human-readable format through seamless integration with ChatGPT, providing valuable insights and context to the end user. The modular nature of the proposed methodology allows integration with various LLMs and XAI techniques, such as LIME and SHAP, as well as different conversational agents such as ChatGPT and Meta's Llama. This allows for full customization, enabling alignment with the requirements of the end user, particularly clinicians. The methodology was tested on two datasets from Twitter and Reddit, demonstrating superior classification accuracy over state-of-the-art models, despite not using additional data for further pre-training. Furthermore, it effectively balances the trade-off between interpretability and accuracy, offering users an explanation of the underlying rationale behind the model decision-making process while maintaining high classification performance. Future work could explore the dynamic nature of individuals' mental states over time, applying our methodology to analyze entire conversational threads, such as those on platforms like WhatsApp. Adapting our methodology for real-time analysis could enable timely alerts to healthcare professionals, enhancing early detection and intervention in critical situations.